%% file: root.tex
\documentclass[letterpaper, 10 pt, journal, twoside]{IEEEtran}

\input{PackagesAndNewCommands}

\bibliography{ref}

\IEEEoverridecommandlockouts
\begin{document}

\title{\LARGE \bf Inferring Foresightedness in Dynamic Noncooperative Games}
\author{Cade Armstrong$^\star$, Ryan Park$^\star$, Xinjie Liu, Kushagra Gupta, and David Fridovich-Keil
\thanks{
Manuscript received: 15 April, 2025; Revised: 25 July, 2025; Accepted: 14 October, 2025. This paper was recommended for publication by Editor Aniket Bera upon evaluation of the Associate Editor and Reviewers’ comments. 

This work was supported by the National Science Foundation under awards 2336840 and 2211548, and by the Army Research Laboratory under Cooperative Agreement Number W911NF-25-2-0021. 
All authors are with the Oden Institute for Computational Sciences and Engineering, University of Texas at Austin, 201 E 24th St, Austin, TX. \tt \{cadearmstrong, ryanjpark, xinjie-liu, kushagrag, dfk\}@utexas.edu

\normalfont
Digital Object Identifier (DOI): see top of this page.
% \smallskip
}
}
\maketitle
\global\csname @topnum\endcsname 0
\global\csname @botnum\endcsname 0
\markboth{IEEE Robotics and Automation Letters. Preprint Version. Accepted October, 2025}
{Armstrong \MakeLowercase{\textit{et al.}}: Inferring Foresightedness in Dynamic Noncooperative Games}

\def\thefootnote{$\star$}
\footnotetext[1]{indicates equal contribution (Corresponding author: Cade Armstrong).}
\def\thefootnote{\arabic{footnote}}
\input{Content/00_Abstract}
\input{Content/01_Introduction}
\input{Content/02_RelatedWork}
\input{Content/03_Background}
\input{Content/04_ProblemStatement}
\input{Content/05_SolutionApproach}
\input{Content/06_ExperimentalResults}
\input{Content/07_DiscussionAndConclusions}

\printbibliography
\end{document}

%% file: PackagesAndNewCommands.tex
\usepackage[natbib=true, style=ieee, maxcitenames=2, mincitenames=1, sorting=none, citetracker, mcite]{biblatex}
\usepackage{amsthm}
\usepackage{amsmath}
\usepackage{cases}
\usepackage{xcolor}
\usepackage[pdftex]{graphicx}
\usepackage{graphicx}
\usepackage{nicefrac}
\usepackage{amssymb}
\usepackage[ruled,vlined,linesnumbered]{algorithm2e}
\usepackage[noend]{algpseudocode}
\usepackage[acronym, shortcuts]{glossaries}
\usepackage{subfigure}
\usepackage{siunitx}
\DeclareGraphicsExtensions{.pdf,.png}
\usepackage{float}
\usepackage[inline]{enumitem}

\usepackage{hyperref}
\hypersetup{hidelinks}
\usepackage[capitalize,noabbrev,nameinlink]{cleveref}
\crefformat{equation}{(#2#1#3)}
\Crefname{algocfline}{Algorithm}{Algorithms}
\Crefname{algocf}{line}{lines}
\Crefname{assumption}{Assumption}{Assumptions}
\crefrangeformat{assumption}{Assumptions~#3#1#4-#5#2#6}

\theoremstyle{plain}

\newtheorem{problem}{Problem}

\newcommand{\edit}[1]{{\textcolor{black}{{#1}}}}

\newcommand{\del}{\nabla}

\newcommand{\expandtop}[3][]{{#2}^1_{#1}, {#2}^2_{#1}, \dots , {#2}^{#3}_{#1}}

\newcommand{\upto}[1]{\{1, 2, ... , #1\}}

\newcommand{\ctrl}{\textbf{u}}
\newcommand{\state}{x}
\newcommand{\fullState}{\textbf{\state}}
\newcommand{\control}{u}
\newcommand{\fullControl}{\textbf{\control}}
\newcommand{\cost}{C}
\newcommand{\runningCost}{J}
\newcommand{\finalTime}{T}
\newcommand{\gameParam}{\theta}
\newcommand{\dynamics}{f}
\newcommand{\obs}{y}
\newcommand{\fullObs}{\textbf{y}}
\newcommand{\ineqConstr}{I}
\newcommand{\eqConstr}{E}
\newcommand{\numPlayers}{N}
\newcommand{\game}{\mathcal{G}}

\newcommand{\initialState}{x_1}
\newcommand{\obsFunc}{h}
\newcommand{\observations}{\textbf{\observation}}
\newcommand{\observation}{y}
\newcommand{\covariance}{\Sigma}
\newcommand{\playerLagrangian}{\mathcal{L}^i}

\newcommand{\gradu}{\nabla_{u^i}}

\newcommand{\ineqMult}{\lambda}
\newcommand{\eqMult}{\mu}

\newcommand{\discountFactor}{\gamma}

\newcommand{\reals}{\mathbb{R}}

\newcommand{\iter}{k}
\newcommand{\micpEqConstr}{c}
\newcommand{\micpIneqConstr}{h}
\newcommand{\probMin}{\mathcal{P}}
\newcommand{\learningrate}{\alpha}

\newcommand{\effectiveFinalTime}{\Tilde{T}}
\newcommand{\distanceMargin}{\delta}
\newcommand{\controlCostWeight}{w_{\textrm{ctrl}}}
\newcommand{\goalCostWeight}{w_{\textrm{goal}}}
\newcommand{\collisionCostWeight}{w_{\textrm{coll}}}
\newcommand{\minDist}{d_{\textrm{min}}}
\newcommand{\micpFreeVar}{r}
\newcommand{\micpNonegVar}{z}
\newcommand{\dimMicpFreeVar}{\eta_{\micpFreeVar}}
\newcommand{\dimMicpNonegVar}{\eta_{\micpNonegVar}}
\newcommand{\paramDim}{k}
\newcommand{\discountFunction}{\Gamma}
\newcommand{\micpVars}{v}
\newcommand{\dPrmDsc}{\nabla_{(\gameParam, \discountFactor)}}
\newcommand{\activeConstInd}{\mathcal{S}}
\newcommand{\inactiveConstr}{\mathcal{I}}
\newcommand{\dt}{\Delta}
\newcommand{\convergenceTolerance}{\epsilon}

\newcommand{\error}{\textrm{error}}
\newcommand{\regConstant}{c}
\newcommand{\crosswalkCost}{\runningCost^{i}_{\mathrm{ped}}}
\newcommand{\lanecenter}{\mathcal{C}}
\newcommand{\lanecenterCostWeight}{w_{\textrm{lane}}}

\newcommand{\ith}{$i^\mathrm{th}$~}

\glsdisablehyper

\newacronym{gps}{GPS}{Global Positioning System}
\newacronym{gne}{GNE}{general Nash equilibrium}
\newacronym{gnep}{GNEP}{general nash equilibrium problem}
\newacronym{golne}{GOLNE}{generalized open-loop Nash equilibrium}
\newacronym{micp}{MiCP}{Mixed Complementarity Problem}
\newacronym{ibr}{IBR}{iterated best response}
\newacronym{kkt}{KKT}{Karush-Kuhn-Tucker}
\newacronym{rl}{RL}{reinforcement learning}
\newacronym{mle}{MLE}{maximum likelihood estimation}

%% file: Content/00_Abstract.tex
\begin{abstract}
Dynamic game theory is an increasingly popular tool for modeling multi-agent, e.g. human-robot, interactions.
Game-theoretic models presume that each agent wishes to minimize a private cost function that depends on others' actions.
These games typically evolve over a fixed time horizon, specifying how far into the future each agent plans.
In practical settings, however,  decision-makers may vary in foresightedness, or how much they care about their current cost in relation to their past and future costs.
We conjecture that quantifying and estimating each agent's foresightedness from online data will enable safer and more efficient interactions with other agents.
To this end, we frame this inference problem as an \emph{inverse} dynamic game.
We consider a specific objective function parametrization that smoothly interpolates myopic and farsighted planning.
Games of this form are readily transformed into parametric mixed complementarity problems; we exploit the directional differentiability of solutions to these problems with respect to their hidden parameters to solve for agents' foresightedness.
We conduct three experiments: one with synthetically generated delivery robot motion, one with real-world data involving people walking, biking, and driving vehicles, and one using high-fidelity simulators.
The results of these experiments demonstrate that explicitly inferring agents' foresightedness enables game-theoretic models to make $\mathbf{33\%}$ more accurate models for agents' behavior.
\end{abstract}

%% file: Content/01_Introduction.tex
\section{Introduction}
Robot planning problems often involve strategic interactions between human and robotic decision makers.
Dynamic game theory captures many of these problems' complexities, such as conflicting goals, coupled strategies, and long-term consequences.
However, for an agent to use game theoretic planning algorithms, it must understand other agents' hidden objectives, or objectives that are not known \emph{a priori}.
To remedy this, we can cast inferring hidden objectives as an inverse game \cite{hadfield2016cooperative,kuleshov2015inverse,molloy2022inverse}. 
In practice, agents can use algorithms to simultaneously infer others' objectives and plan trajectories that seamlessly interact with one another over time. 

Existing inverse game solvers often assume that decision makers care equally about each moment in time, \edit{but in practice, humans and robots weigh future costs differently than present costs.}
For example, in an urban driving game modeling a busy intersection, a farsighted driver may slow down further in advance of the intersection than a short-sighted, or myopic, driver would.
Robots incorrectly predicting the myopic driver's foresight may take prematurely evasive actions, resulting in unexpected and unnecessarily risky behavior.

\begin{figure}
    \centering
    \includegraphics[scale=0.55]{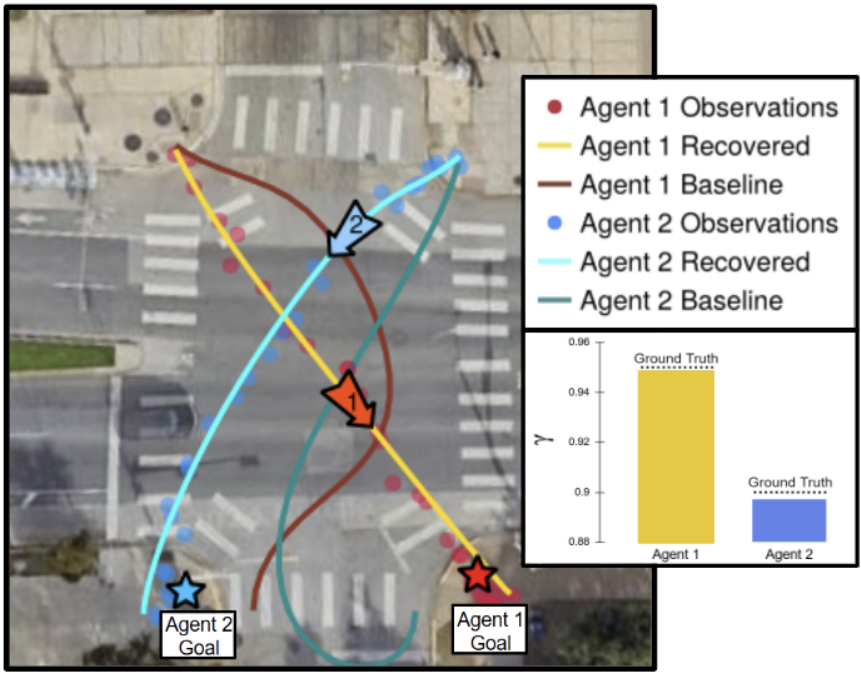}
    \caption{Representative example showcasing how the proposed formulation improves our ability to model agents' behavior in noncooperative interactions. 
    Here, two simulated \edit{delivery robots} are crossing an intersection, and by correctly inferring their degree of foresightedness (i.e., the $\discountFactor$ parameter in the inset), our method recovers their trajectories than a baseline approach.
    \vspace{-0.5cm}
    }
    \label{fig:trial_image}
\end{figure}

This paper aims to address the gap between the assumptions of existing game theoretic algorithms and decision makers' foresightedness.
In particular, existing methods typically model each agent's objective as a sum of cost functions which can depend upon all agents' states and actions.
In this work, we consider a time-discounted cost formulation, which smoothly interpolates foresight for each agent.
Additionally, we develop a gradient-based algorithm for inverting the corresponding game and identifying each agent's degree of foresightedness.
Our specific contributions are: \edit{1) an explicitly foresighted game formulation, 2) an efficient method to infer each agent's degree of foresightedness from data collected online, and 3) and a series of simulated and real-world experiments that demonstrate the benefits of modeling agents' foresight.}

\edit{In particular, we conduct one simplified experiment with two myopic delivery robots (presented in \cref{fig:trial_image}) and two simulated experiments with real-world data (presented in \cref{fig:InD_vis} and \cref{fig:waymax}).}
Results demonstrate that our method recovers more accurate models of foresighted agents than existing models, as well as being more robust to noise in partially observable settings.
Specifically, our method reduces trajectory error by over $33\%$ on average in comparison to cutting-edge baseline methods.

%% file: Content/02_RelatedWork.tex
\section{Related Works}
\subsection{Noncooperative Dynamic Games}
An \(\numPlayers\)-agent dynamic game models each agent's choice of strategy in terms of a optimization problem, whose objective and/or constraints can depend upon other agents' strategies.
Dynamic games admit different solutions depending on the information available to each agent at every time step \cite{BasarOlsder}. 
In this work, we consider the \emph{Nash} equilibrium concept in the setting of \emph{open-loop} strategies, i.e., we presume that at each time $t$ agents only have information about the initial state of the game. 
Early works in this setting focused primarily on problems with convex, quadratic costs and linear game dynamics, without any additional constraints \cite{starr1969nonzero, BasarOlsder, engwerda1998open, engwerda1998computational}. 
However, finding a Nash equilibrium in more general settings is often intractable \cite{daskalakis2009complexity}. Consequently, many recent works develop iterative trajectory optimization methods to find \emph{approximate, local} open-loop Nash equilibria in games with nonlinear costs and complex constraint structures. 
Such approximations satisfy first-order optimality conditions of the underlying optimization problem for each agent, and second-order conditions for local optimality must generally be checked \emph{a posteriori}. 
\edit{Some well-known methods include root-finding techniques for the underlying joint $\numPlayers$-agent \ac{kkt} system \cite{cleac2019algames, di2020first,zhu2023sequential} and iterative best-response algorithms to identify local Nash equilibria \cite{spica2020real,wang2019game}.}

\subsection{Inverse Dynamic Games}

Inverse dynamic games seek to identify unknown aspects of game models, e.g., agents' intentions and constraints, from observed interactions. 
Early approaches minimize the residual of the KKT system to identify possible game parameter values~\cite{ROTHFU201714909,awasthi2020inverse,inga2019inverse}.
However, this type of approach requires \emph{full} state-action demonstration to evaluate the residuals, posing a challenge for online estimation tasks, where not all agents' states and actions are available. 
Worse, the assumption that the demonstrations satisfy first-order optimality conditions means that such residual minimization approaches can perform poorly in the presence of observation noise~\cite{peters2023ijrr}. 
To mitigate these issues, recent work formulates inverse game problems in terms of \ac{mle}, and explicitly impose first-order optimality conditions as constraints~\cite{peters2023ijrr,liu2023learningplaytrajectorygames,li2023cost,hu2024think}. 
In particular, our work builds directly upon this line of work, and specifically extends the solution approach of \cite{liu2023learningplaytrajectorygames}, which develops a differentiable game solver that enables inverse game problems to be solved via gradient descent on the unknown game parameters.

While \ac{mle} approaches to inverse games demonstrate strong performance in various scenarios, they are fundamentally limited to providing point estimates of unknown parameters.
To this end, another line of work seeks to infer full Bayesian posterior \emph{distributions} for the unknown game parameters, e.g., via particle filtering~\cite{peters2020accommodating} and unscented Kalman filtering~\cite{le2021lucidgames}. 
More recently, work in \cite{liu2024autoencodingbayesianinversegames} proposes to embed a differentiable Nash solver into generative models to solve variational inference problems. 
However, high computational costs hinder these approaches from scaling to larger systems. 

Finally, we note that inverse games have also been formulated in maximum-entropy settings~\cite{mehr2023maxent}, inferring opinion dynamics model~\cite{hu2024think}, and in combination with neural network components~\cite{diehl2023energy}. 
\edit{
\subsection{Non-Game Theoretic Multi-Agent Modeling}
Many non-game theoretic approaches have been proposed for modeling multi-agent robotic systems.
Social forces have been used to model multi-agent interactions, where agent objectives are introduced through virtual forces (e.g., spring forces to attract an agent) \cite{social_forces,social_swarm}.
Potential fields have been used similarly; instead of virtual forces to encode goals and obstacles, the method uses potential fields to repel and attract agents \cite{potential_fields}.
Relevant works model multi-agent behavior by encoding agent objectives in velocity space rather than position space with velocity obstacles \cite{velocity_obstacles, vel_obs_goals}.
Additionally, recent work applies machine learning to predict human driving and pedestrian behavior. 
Maximum entropy inverse reinforcement learning \cite{ziebart2008maximum,pedestrian_learning} and history-dependent LSTM-based models \cite{deo2018convolutionalsocialpoolingvehicle,ridel2019understandingpedestrianvehicleinteractionsvehicle} are examples of learning methods that can forecast vehicle and pedestrian trajectories. 
Finally, recent works use generative trajectory models to predict pedestrian behavior \cite{trajectron,trajectron++}.
Unlike game-theoretic methods, these methods do not explicitly reason about coupling effects in agents' decision-making.
Additionally, it should be noted that our approach exhibits the interpretability of classical methods, such as social force models, but can flexibly accommodate arbitrary specifications of agents' objective functions, similar to inverse reinforcement learning methods. 
}

%% file: Content/03_Background.tex
\section{Background} 
For notational convenience, lowercase bold variables refer to aggregations over time and the omission of agent indices denotes aggregations of all agents. 
In addition, agent indices are always written as superscripts while time indices are always written as subscripts.

\subsection{Dynamic Games}
A dynamic game  \cite{BasarOlsder} is characterized by $\numPlayers$ agents, with the \ith agent's control input denoted $\control^i_t \in \mathbb{R}^{m^i}$ \edit{for all discrete times $t \in [\finalTime]:=\upto{\finalTime}$ and a joint state variable $\state_t \in \mathbb{R}^n$ following given dynamics $\state_{t+1} = \dynamics_t(\state_t, \expandtop[t]{\control}{\numPlayers})$.}
Each agent has a cost function
\begin{align}
\runningCost^i := \sum_{t=1}^\finalTime \discountFunction^i(t; \discountFactor^i) \cost^i(\state_t, \control^i_t, \control^{\neg i}_t; \gameParam^i),
\label{eq:playerCost}
\end{align}
which depends upon the state and its own actions, as well as others' actions $u^{\neg i}_t$, hidden parameters $\gameParam^i \in \reals^\paramDim$, and \emph{discount factor} $\discountFactor^i$.
At each time $t$, agent $i$'s cost is comprised of parametrized function $\cost^i(\cdot)$ and scaled by a parametrized \textit{discounting function} $\discountFunction^i(t;\discountFactor^i)$, quantifying the importance of agent $i$'s cost at time $t$.

In principle, the formulation of the game in \cref{eq:playerCost} is general enough to handle arbitrary functions $\discountFunction^i$, which are twice differentiable in $\discountFactor^i$.
In this work, we assume that this discounting function is an exponential, i.e. $\discountFunction^i(t; \discountFactor^i) = (\discountFactor^i)^t$, where $\discountFactor^i \in (0, \infty]$.

We shall refer to all agents' discount factors as a vector
$\discountFactor = (\discountFactor^1, \dots, \discountFactor^\numPlayers) \in [0, \infty]^\numPlayers$ for convenience.
Observe that the function $\discountFunction^i(t; \discountFactor^i)$ rapidly approaches $0$  as $t \to \finalTime$ when $\gamma^i < 1$. 
Thus, the \ith agent's actions do not depend upon costs incurred past some effective time horizon $\effectiveFinalTime^i \in [\finalTime]$.

Thus, we refer to the sequence of game states as $\fullState = (\state_1^{\top}, \state_2^{\top}, \dots, \state_\finalTime^{\top})^\top$, agent $i$'s control sequence as $\fullControl^i = (\control^{i, \top}_1, \control^{i, \top}_2, \dots, \control^{i, \top}_\finalTime)^\top$, the sequence of all agents' actions as $\fullControl = (\fullControl^{1, \top}, \fullControl^{2, \top}, \dots, \fullControl^{\numPlayers, \top})^\top$, and all agents' hidden parameters as $\gameParam = (\gameParam^{1, \top}, \gameParam^{2, \top}, \dots, \gameParam^{\numPlayers, \top})^\top$.
Then, we can write each cost function as $\cost^i_t(\state_t, \control^i_t, \control^{\neg i}_t; \gameParam^i) = \cost^i_t(\state_t, \control_t; \gameParam^i)$ and thus agent $i$'s overall cost is $\runningCost^i(\fullState, \fullControl; \discountFactor^i, \gameParam^i)$ with a slight abuse of notation.
In general, we may also assign each agent a set of inequality constraints $I^i(\fullState, \fullControl; \gameParam^i) \ge 0$ and a set of equality constraints $E^i(\fullState, \fullControl; \gameParam^i) = 0$.
Finally, we can define a game as a tuple:
\begin{equation}
\label{eq:game_definition}
\begin{aligned}
\game(\gameParam, \discountFactor) = (&\{\runningCost^i(\cdot;\discountFactor^i, \gameParam^i)\}_{i \in [N]},\{\ineqConstr^i(\cdot; \gameParam^i)\}_{i \in [N]}, \\
&\{\eqConstr^i(\cdot; \gameParam^i)\}_{i \in [N]},\, \initialState,\, \finalTime,\, \numPlayers). 
\end{aligned}
\end{equation}
Note that \cref{eq:game_definition} does not include the dynamics $\dynamics_t(\cdot)$ explicitly; this is because the dynamics generate $(\finalTime - 1)$ equality constraints of the form $\state_{t+1} - \dynamics_t(\state_t, \expandtop[t]{\control}{\numPlayers}) = 0$ for each of the $\numPlayers$ agents.
For clarity, we assume that $\eqConstr^i(\cdot)$ contains these terms. 

\subsection{Generalized Open-Loop Nash Equilibria}
\label{sec:golne}
For a given set of parameters $\gameParam$ and discount factors $\discountFactor$, a \acrfull{golne} of the game $\game(\gameParam, \discountFactor)$ in \cref{eq:game_definition} is given by a point $(\fullState^*, \fullControl^*)$  which jointly solves the following coupled optimization problems: 
\begin{subequations}
\label{eq:nash}
\begin{numcases}{\forall i \in [\numPlayers]}
    \label{eq:coupled_optimization_problems}
    \min_{\fullState, \fullControl^i} ~ \sum_{t=1}^\finalTime \discountFunction^i(t; \discountFactor^i) \cost^i(\state_t, \control^i_t, \control^{\neg i}_t; \gameParam^i)\\
    \text{s.t. }
    \label{eq:constraints}
    \eqConstr^i(\fullState, \fullControl; \gameParam^i) = 0 \\
    \hspace{0.58cm} \ineqConstr^i(\fullState, \fullControl; \gameParam^i) \ge 0.  \notag
\end{numcases}
\end{subequations}
The strategies $\fullControl^*=(\ctrl^{1*}, \ctrl^{2*}, ... , \ctrl^{N*})$  have the property that $\runningCost^i(\fullState, \ctrl^i, \ctrl^{\neg i*}; \gameParam^i, \discountFactor^i) \ge \runningCost^i(\fullState^*, \fullControl^*; \gameParam^i, \discountFactor^i)$ for all feasible $\fullControl^i$ and $\fullState$, for all agents $i \in [\numPlayers]$. 

\subsection{\texorpdfstring{\acf{micp}}{Mixed Complementarity Problem (MiCP)}} \label{sec:MiCP}

Finding a generalized Nash equilibrium is computationally intractable \cite{papadimitriou2007complexity}; therefore, it is common to relax the condition in \cref{sec:golne} to hold only within an open neighborhood of the point $(\fullState^*, \fullControl^*)$ \cite{mazumdar2019finding,mazumdar2020gradient,heusel2017gans,di2020first,zhu2023sequential}.
Such points are called local Nash equilibria \cite{ratliff2016characterization}.
Practically, such points can be identified by solving agents' first order necessary conditions, which constitute a \acf{micp} \cite{facchinei_pang}.
\edit{An \acrshort{micp} is defined by
decision variables $\micpFreeVar \in \reals^{\dimMicpFreeVar} , \micpNonegVar \in \reals^{\dimMicpNonegVar}$, as well as functions $\micpEqConstr(\micpFreeVar, \micpNonegVar)$ and $\micpIneqConstr(\micpFreeVar, \micpNonegVar)$, such that $\micpEqConstr(\micpFreeVar, \micpNonegVar) = 0 \text{ and } 0 \le \micpNonegVar \perp \micpIneqConstr(\micpFreeVar, \micpNonegVar) \ge 0.$}
\acsp{micp} can often be solved efficiently via off-the-shelf solvers such as PATH \cite{dirkse1995path}.

%% file: Content/04_ProblemStatement.tex
\section{Problem Statement}
We presume that an observer has obtained noisy sensor measurements of the game state over time, and denote these observations $\fullObs = \big[\obs_1(\state_1), \obs(\state_2), \obs_3(\state_3),  \hdots,\obs_T(\state_\finalTime)\big]$.
We assume observations are drawn independently from Gaussian models $\obs_t(\state_t) \sim \mathcal{N}\big(\obsFunc_t(\state_t), \covariance_t\big)$ with known covariances $\covariance_t$, where $\obsFunc_t(\state_t)$ describes the expected output of the sensor at time $t$.

Given measurements $\observations$, we seek to identify game parameters $\hat{\gameParam}, \hat{\discountFactor}$ that maximizes the likelihood to have observed $\observations$ from a \acs{golne} of $\game(\hat{\gameParam}, 
\hat{\discountFactor})$.
Equivalently, one can minimize covariance-weighted deviations from expected measurements, which we denote $\probMin$.
\begin{problem} \label{prb:InverseProblemStatement}
\textbf{Inverse Game Problem:} Given a sequence of observations $\observations$, find parameters $\hat{\gameParam}$ which solve
\end{problem}
\begin{subequations}
\vspace{-0.5cm}
    \begin{alignat}{2}
        &\min_{\fullState, \fullControl, \gameParam, \discountFactor}~ && \overbrace{\sum_{t = 1}^\finalTime (\obsFunc_t(\state_t) - \observation_t)^\top \covariance^{-1}_t  (\obsFunc_t(\state_t) - \observation_t)}^{\probMin\left(\fullState(\gameParam, \discountFactor)\right)} \label{eq:cov_distance_prob}  \\
        &\text{s.t.}~ &&(\fullState, \fullControl)\text{ is a \acs{golne} of }\mathcal{G}(\gameParam, \discountFactor). \label{eq:gne_constr}
    \end{alignat}
\end{subequations}

Intuitively, \cref{prb:InverseProblemStatement} can be separated into an outer \emph{inverse} problem and an inner \emph{forward} problem. 
The forward \acs{golne} problem  \cref{eq:gne_constr} is parametrized by the game parameters $\gameParam$ and discount factors $\discountFactor$, which are decision variables in the inverse problem \cref{eq:cov_distance_prob}.
This formulation is general enough to handle partial state observations, making it amenable to realistic observation models such as cameras or \acrshort{gps}, in principle.
However, solving \cref{eq:cov_distance_prob} entails computational challenges.
In particular, the \acs{kkt} conditions of \cref{eq:gne_constr} are certainly nonlinear in $\discountFactor$, making the overall problem non-convex.
In addition, \cref{eq:gne_constr} may contain inequality constraints, which imply that its \acs{kkt} conditions involve complementarity conditions which make the overall problem non-smooth.

%% file: Content/05_SolutionApproach.tex
\section{Solution Approach}
\label{sec:solution_approach}
In this section, we present a constrained gradient descent algorithm for identifying unknown parameters $\gameParam, \discountFactor$ in \cref{prb:InverseProblemStatement}. 
This technique will require us to take derivatives of solutions $(\fullState, \fullControl)$ to \cref{eq:gne_constr} with respect to parameters $(\gameParam, \discountFactor)$. 
Therefore, we begin with a discussion of transforming a \acrshort{golne} into an \acrshort{micp}, whose solutions are directionally differentiable with respect to problem parameters \cite[Ch. 5]{facchinei_pang}.

\subsection{Equilibrium Constraint as an \texorpdfstring{\acs{micp}}{(MiCP)}}
\label{sec:gnep_as_micp}
In this subsection, we show how to convert a \acs{golne} into a \acs{micp} which encodes its first-order necessary conditions.
Similar techniques have been used in \cite{mombaur2010human,englert2017inverse,peters2023online,liu2023learningplaytrajectorygames} to solve inverse optimal control and game problems.

We start with writing the agents' first-order necessary conditions. 
First, we introduce Lagrange multipliers $\ineqMult^i$ and $\eqMult^i$ for the \ith agent's inequality and equality constraints, respectively, and write its Lagrangian as
\begin{multline}
    \playerLagrangian(\fullState, \fullControl, \ineqMult^i, \eqMult^i; \gameParam^i, \discountFactor^i) = \cost^i(\fullState, \fullControl; \gameParam^i, \discountFactor^i) - \\ 
    \ineqMult^{i,\top} I^i(\fullState, \fullControl; \gameParam^i) - \eqMult^{i,\top} E^i(\fullState, \fullControl; \gameParam^i).
\end{multline}
Then, when the gradients of the constraints are linearly independent at a candidate solution point (i.e., the linear independence constraint qualification is satisfied \cite[Chapter 3.2]{facchinei_pang}),  the following \acrfull{kkt} conditions must hold for each agent $i$:  
\edit{\begin{subequations}
\label{eq:kkt}
\begin{align}
    \nabla_{\fullState} \playerLagrangian = 0, 
    \gradu \playerLagrangian = 0, 
    E^i &= 0\\
    0 \leq \ineqMult^i \perp  I^i(\fullState, \fullControl; \gameParam^i) &\geq 0.
\end{align}
\end{subequations}}
We can then structure agents' joint \acrshort{kkt} conditions as a parametric \edit{\acs{micp}}, in which the primal and dual variables are concatenated as
\edit{
$\micpFreeVar = (\fullState^\top, \fullControl^{\top}, \eqMult^{1,\top}, \eqMult^{2,\top}, \dots, \eqMult^{N,\top})^\top$ and
$\micpNonegVar = (\ineqMult^{1, \top}, \ineqMult^{2,\top}, \dots, \ineqMult^{N,\top})^\top$.
}
For brevity, define $\micpVars = (\micpFreeVar^\top, \micpNonegVar^\top)^\top$.
Then, with a slight abuse of notation, the parameterized \acrshort{micp} for each agent can be written as described in \edit{\cref{sec:MiCP}}, where 
\edit{
$c(\micpVars; \gameParam, \discountFactor) = 
    [
        \left(\nabla_{\fullState} L^i\right)^\top_{i \in [\numPlayers]},
        \left(\nabla_{\fullControl^i} L^i\right)^\top_{i \in [\numPlayers]},
        \left(E^i\right)^\top_{i \in [\numPlayers]}
    ]^\top$ and $
    h(\micpVars; \gameParam, \discountFactor) = [
        \bigl(
        I^i
        \bigr)_{i \in [\numPlayers]}
    ]^\top
$}.
For brevity, we will define $F(\micpVars; \gameParam, \discountFactor) = [c(\cdot)^\top, h(\cdot)^\top]^\top$. 
For additional details on connections between \acrshort{golne}s and \acsp{micp} in the context of open loop dynamic games, we direct the reader to \cite{facchinei_pang, liu2023learningplaytrajectorygames}.

\subsection{Optimizing Parameters with Gradient Descent} \label{sec:inv_grad_descent}

In this section, we present our algorithm for solving \cref{prb:InverseProblemStatement}.
We first replace \cref{eq:gne_constr} with the \acs{kkt} conditions in \cref{eq:kkt}, and transcribe them into an \acs{micp} according to \cref{sec:gnep_as_micp}.
Then, we compute the total derivative of objective function $\probMin$ with respect to parameters $(\gameParam, \discountFactor)$, and update their values accordingly. 
These gradients can be computed via the chain rule:\vspace{-0.2cm}
\begin{align}
\label{eq:probGrad}
    \nabla_{(\gameParam, \discountFactor)} \probMin(\fullState(\gameParam, \discountFactor)) = (\nabla_{(\gameParam, \discountFactor)}\micpVars)^\top (\nabla_{\micpVars}\fullState)^\top (\nabla_{\fullState}\probMin).
\end{align}
The only non-trivial term in \cref{eq:probGrad} is $\nabla_{(\gameParam, \discountFactor)}\micpVars$.
Next, we show how to take directional derivatives of \edit{\acsp{micp}} at a solution $\micpVars^*$ where strictly complementarity holds. 

First, we must consider the complementarity constraints on $\micpNonegVar$ and $h(\cdot)$.
To this end, we construct an index set $\inactiveConstr$ which records all inactive inequality constraint dimensions in $h(\micpVars; \gameParam, \discountFactor)$.
Indexing $F$ at elements of this set yields a vector $[F]_{\inactiveConstr}$.
\edit{From \cref{sec:MiCP}, }we know that these inactive inequalities $[F]_{\inactiveConstr}$ are strictly positive, and the Lagrange multipliers associated with these constraints are exactly $0$.
Then, presuming the continuity of $F$, small changes in $(\gameParam, \discountFactor)$ preserve the positivity of elements of $[F]_{\inactiveConstr}$ and force the corresponding Lagrange multipliers in $[\micpVars]_\inactiveConstr$ to remain $0$.
Thus, we find that $\dPrmDsc [\micpVars]_\inactiveConstr = 0$.

Consider the remaining constraints from \cref{eq:kkt}, which must be active due to strict complementarity; denote the indices corresponding to these constraint as $\activeConstInd$.
We can use the implicit function theorem and stationarity of $F$ with respect to $\micpVars$ to write:
\vspace{-0.2cm}
\begin{subequations}
\label{eq:inactive_constraint_handling}
    \begin{align}
    \label{eq:active_constraint_total_d}
    0 &= \dPrmDsc [F]_{\activeConstInd}  = \dPrmDsc [F]_{\activeConstInd} + (\nabla_{\micpVars} [F]_{\activeConstInd})(\dPrmDsc \micpVars)\\
    &\implies \dPrmDsc \micpVars = -(\nabla_{\micpVars} [F]_{\activeConstInd})^{-1} \dPrmDsc [F]_{\activeConstInd}. 
    \end{align}
    \end{subequations}
Then, when $\nabla_{\micpVars} [F]_{\activeConstInd}$ is invertible, we can find exact values of $\dPrmDsc \micpVars$.
For a more complete treatment, including a discussion of weak complementarity, readers are encouraged to consult \cite{dontchev2009implicit, facchinei_pang, liu2023learningplaytrajectorygames}.

Thus equipped, we iteratively update our estimate of $(\gameParam, \discountFactor)$ according to the gradient of \cref{eq:cov_distance_prob}---incorporating the aforementioned implicit derivatives as needed.
\cref{alg:1} terminates after a maximum number of iterations (i.e., when $k > K$) or when parameters have converged (i.e., $\|\gameParam_{k+1} - \gameParam_k\|_2 \le \convergenceTolerance$). 
In our experiments, $K = 500$ and $\convergenceTolerance = 10^{-4}$. At each step, also note that we project the discount factor $\discountFactor$ onto the set $[0, \infty)^\numPlayers$ to ensure feasibility. 

\begin{algorithm}[ht!]
\DontPrintSemicolon
\caption{Foresight-aware inverse game} 
\label{alg:1}
{
\textbf{Hyper-parameters:} Learning rate $\learningrate$\\
\textbf{Input:} initial $\gameParam$, initial $\discountFactor$, observations $\observations$\\
$\gameParam_0 \gets \gameParam$ \\
$\discountFactor_0 \gets \discountFactor$ \\
$\iter \gets 0$ \\
\While{not converged}{
     $\micpVars_{\iter}
    \gets \operatorname{solveInnerMCP}(\gameParam_{\iter}, \discountFactor_{\iter})$ \label{line:solve-micp}  \Comment{\edit{\cref{sec:gnep_as_micp}}} \\
     $\del_{(\gameParam_\iter, \discountFactor_\iter)} \probMin \gets \edit{\operatorname{calcGrad}}(\micpVars_k, \gameParam_k, \discountFactor_k)$ \Comment{\edit{\cref{sec:inv_grad_descent}}} \\
     $\gameParam_{\iter+1} \gets \gameParam_\iter - \del_{\gameParam_\iter} \probMin \cdot \learningrate$ \\    
     $\discountFactor_{\iter+1} \gets \max (0, \discountFactor_\iter - \del_{\discountFactor_\iter} \probMin \cdot \learningrate )$ \\
     $\iter \gets \iter + 1$ \\
} 
\textbf{return} $(\gameParam_\iter, \discountFactor_\iter, \fullState, \fullControl)$
}
\end{algorithm}

%% file: Content/06_ExperimentalResults.tex
\section{Experiments}
In this section, we test the performance and robustness of the proposed approach in \cref{sec:solution_approach} and compare it to baseline inverse game approaches in simulated and real-world experiments.
Results show that \cref{alg:1} performs better than an existing inverse game formulation across various performance metrics. 
\subsection{Example Game - Crosswalk}
\label{sec:experimental_setup}
% \begin{figure}
%     \subfigure[Fully-observed case\label{fig:fully_obs_heatmap}]{
%     \includegraphics[width = 0.465\columnwidth]{images/fo_hm_0.6_0.6.pdf}
%     }
%     \subfigure[Partially-observed case\label{fig:partially_obs.png}]{
%     \includegraphics[width = 0.465\columnwidth]{images/po_hm_0.6_0.6.pdf}
%     }
%     % \hfill
%     \caption{ Here we plot a heatmap of $\probMin$ from \cref{prb:InverseProblemStatement} in a (a) fully observable and (b) partially observable setting.  
%     In both cases, both player's $\discountFactor$ are varied while holding $\gameParam$ constant at the ground truth values and the resulting cost is scaled to be in between $[0, 1]$.
%     }
%     \label{fig:heatmaps}
% \end{figure}
Our first experiment involves $\numPlayers = 2$ pedestrians navigating a crosswalk at an intersection on the UT Austin campus at W Dean Keaton St. and Whitis Ave.
This intersection becomes safe for all pedestrian crossing at once, encouraging students to cross diagonally and interact with one another.
We simulate two short-sighted \edit{delivery robots} who begin at the top right and left corners, respectively, and their goals are to reach the bottom left and right corners, respectively.

We model \edit{agents as delivery robots wishing to minimize the distance to their goal and their control effort}, such that the \ith agent's cost can be written as:
% \vspace{-0.2cm}
\begin{align}
    \runningCost^i_{\mathrm{stud}} = \sum_{t=0}^\finalTime(\discountFactor^i)^t ~ \bigr[ \goalCostWeight^i ~\|p^i_t  -  \gameParam^i_{\textrm{goal}}\|^2_2  + \controlCostWeight^i~\|u^i_t\|^2_2].
\end{align}
For brevity, we define $p^i_t \in \reals^2$ to denote 
% the part of $\state_t$ which corresponds to 
agent $i$'s position at time $t$, weigh the goal and control cost terms with $w^i_\textrm{goal}$ and $w^i_\textrm{ctrl}$, and parameterize the game with the \ith agent's target position $\gameParam^i_{\textrm{goal}} \in \reals^2$. 
We model the \edit{robots} as point masses with bounds on the velocity and control inputs, assume a time discretization of $\dt = \SI{0.1}{\second}$, and play the game over $\finalTime = 25$ time steps. 
Explicitly, the game's state at time $t$ is composed as $\state_t = [\state^{1\top}_t, \state^{2\top}_t]^\top$, and 
\edit{
each agent's physical dynamics are modeled with a standard double integrator linear system.
}
% \begin{align*}
%     x_{t+1}^i = 
%     \begin{bmatrix}
%         1 & 0 & \dt & 0\\
%         0 & 1 & 0 & \dt\\
%         0 & 0 & 1 & 0\\
%         0 & 0 & 0 & 1
%     \end{bmatrix}
%     \underbrace{\begin{bmatrix}
%         p_{x,t}^i\\
%         p_{y,t}^i\\
%         v_{x,t}^i\\
%         v_{y,t}^i
%     \end{bmatrix}}_{\state_t^i}
%     + 
%     \begin{bmatrix}
%         \nicefrac{\dt^2}{2} & 0\\
%         0 & \nicefrac{\dt^2}{2}\\
%         \dt & 0\\
%         0 & \dt
%     \end{bmatrix}
%     \underbrace{\begin{bmatrix}
%         a_{x,t}^i\\
%         a_{y,t}^i
%     \end{bmatrix}}_{\control_t^i}
% \end{align*} \cade{just say double integrator dynamics}

Additionally, the \ith agent has a collision avoidance constraint defined as
\begin{align}
    \label{eq:simInequalities}
    I^i(\cdot) = \Big[\big(\|p^i_t - p^{\neg i}_t\|^2_2 -\distanceMargin \cdot \minDist\big)_{t\in[\finalTime]}\Big] ~\ge 0,
\end{align}
where $\minDist$ denotes the minimum allowed distance between the agents and $\distanceMargin > 1$ is a safety margin.
% In our testing, we found that \cref{alg:1} converged more reliably when these constraints are implemented as large penalties in each agent's cost function.
\edit{However, by moving this constraint to each agent's objective, we can avoid assigning a new Lagrange multiplier to each pair of agents.
This avoids quadratic scaling in the number of \acrshort{micp} variables with respect to number of agents.}
Therefore, we model \ith agent's objective function in the inverse game as $\runningCost^i_{\mathrm{ped}} = \runningCost^i_{\mathrm{stud}} + \runningCost^i_{\mathrm{coll}}$, where
% \vspace{-0.2cm}
\begin{multline}
    \runningCost^i_{\mathrm{coll}} = \sum_{t=0}^\finalTime(\discountFactor^i)^t ~ \collisionCostWeight^i ~ \max (0, \distanceMargin \cdot \minDist - \|p^i_t - p^{\neg i}_t\|^2_2 \bigl),
    \label{eq:simCost}
\end{multline}
in which $\collisionCostWeight^i$ is a weighting parameter quantifying the importance of collision avoidance for the \ith agent.
\subsection{Example Game - Intersection}
\label{sec:InD_setup}
\begin{figure}
    \centering
    \includegraphics[width = 0.9\columnwidth]{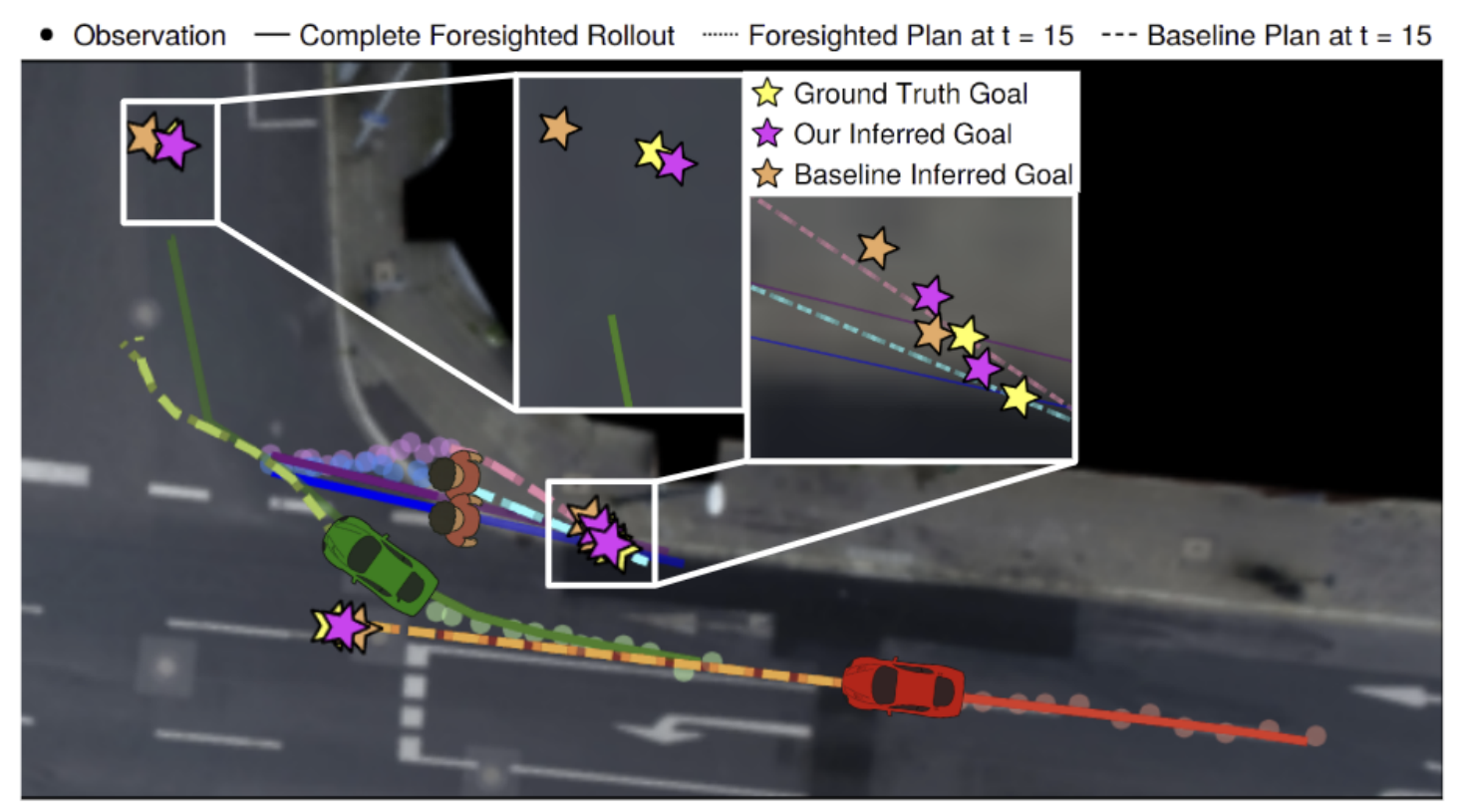}
    \caption{Intersection from the InD dataset \cite{inDdataset}, overlaid with trajectories for all four agents generated in a receding horizon fashion using our method and the baseline. Note that we reduced the opacity of the complete foresighted rollout after $t$ = 15 for visual clarity.
    % \vspace{-0.5cm}
    }
    \label{fig:InD_vis}
\end{figure}
Our second experiment involves $\numPlayers = 4$ agents navigating an intersection: two pedestrians crossing a crosswalk, a car attempting to turn while the pedestrians are crossing, and a car attempting to go straight while behind the turning car.
We employ real-world data of this interaction from the Intersection Drone (InD) dataset, captured by leveLXData via a drone hovering over the intersection found in Aachen, Germany, seen in \cref{fig:InD_vis} \cite{inDdataset}.
Specifically, the interaction takes place over 162 frames at 25 frames per second. 
The data was downsampled by a factor of 6, resulting in a game horizon of $\finalTime = 28$.

The setup for our model in this example is very similar to that discussed in \cref{sec:experimental_setup}. 
To begin, all agents have inequality constraints \cref{eq:simInequalities}. 

We model the cars' physical dynamics as bicycles with bounds on the control inputs. 
Explicitly, the game's state at time $t$ is composed as $\state_t = [\state^{1\top}_t, \state^{2\top}_t]^\top$ \edit{in which $\state_t^i = [p_{x,t}^i, p_{y,t}^i, v_{t}^i, \psi_{t}^i]^\top$,} and for the \ith agent:
\begin{align*}
    {x}_{t+1}^i = 
    &[
        p_{x,t}^i,~
        p_{y,t}^i,~
        v_{t}^i,~
        \psi_{t}^i
    ]^\top+ \\
    \dt
    &[
        v^i_t \cos(\psi_t^i),~
        v^i_t  \sin(\psi_t^i),~
        a_t^i,~
        v_t^i  \tan(\phi_t^i)/l
    ]^\top
\end{align*}
% \ryan{say bicycle dynamics and cite? try not to. Can try breaking into two 2x2 side by side equations. axe underbrace. write state inline.}
with control input $\control_t^i = [a_t^i, \phi_t^i]^\top$ representing longitudinal acceleration and steering angle ($v^i_t$ is longitudinal velocity, $\psi^i_t$ is heading, and $l$ is wheelbase).
Meanwhile, the pedestrians have the same dynamics as discussed in \cref{sec:experimental_setup}.

Depending on whether the \ith agent is a car or pedestrian, we model the agent's objective function slightly differently. 
For the pedestrians, their cost is exactly \edit{$\crosswalkCost$ while cars also minimize} the distance to their lane centers. 
We model the \ith agent's lane center as a set, $\lanecenter^i = \{(c_{x,t}^i,c_{y,t}^i)\}_{t = 0}^\finalTime$ generated from a polynomial curve fit to lane center points. 
We then can write the cost as:
\begin{align}
    \runningCost^i_{\mathrm{car}} = \crosswalkCost + \sum_{t=0}^\finalTime(\discountFactor^i)^t ~ \lanecenterCostWeight^i~ \|p^i_t-c^i_{t}\|^2_2,
\end{align}
where $\lanecenterCostWeight^i$ weighs the lane center cost term and $c^i_t$ refers to the $t^{\mathrm{th}}$ element of set $\lanecenter^i$. 

Lastly, the environment in this example has defined regions in which the agents are unable to traverse (e.g. buildings). 
Thus, we create inequality constraints such that the agents are constrained to the roads/sidewalks. 
To do so, we model corners as circles and use sigmoids for the sides of the road.
\edit{
\subsection{Example Game - Waymax Simulation}
\begin{figure}
    \centering
    \includegraphics[width=\columnwidth]{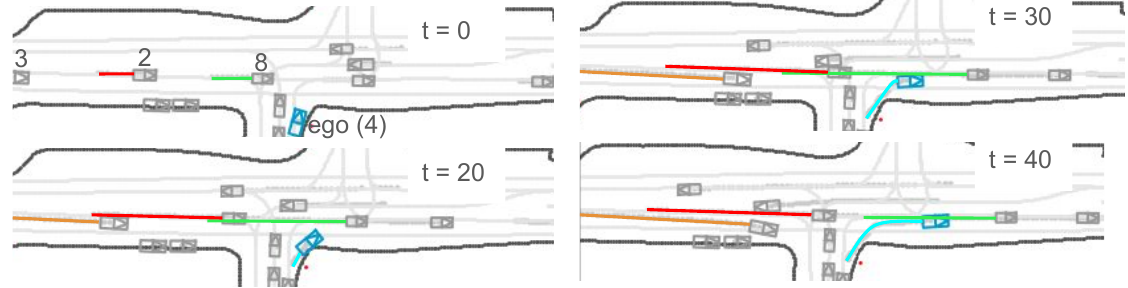}
    \caption{\edit{
    Snapshots from receding horizon inference and planning with the Waymax simulator, for a game involving the blue, green, yellow, and red agents.
    The ego robot (blue) employs our method, while the other cars are controlled by the Waymax simulator.
    }
    \vspace{-0.5cm}}
    \label{fig:waymax}
\end{figure}
Our last experiment involves $N = 4$ cars navigating an intersection, one of which is a robot attempting to make a left turn (cf. \cref{fig:waymax}).
We employ the Waymax simulator, which has been built atop the Waymo Open Motion Dataset \cite{waymax}, to simulate the humans' responses to the ego's actions. 
The ego robot infers the humans' objectives and foresightedness and plans to reach its goal safely in a receding horizon setup.
Specifically, the scenario takes place over a horizon of $T = 50$ steps with a time discretization of $\dt = \SI{0.1}{\second}$ and a planning horizon $T_{\mathrm{plan}} = 11$ steps.}

\edit{Like in \cref{sec:InD_setup}, we model the cars' physical dynamics as bicycles with bounds on the control inputs.
However, we modify the costs from the previous examples by: 1) simplifying the lane center term to only consider $p_{x,t}^i$ as the road is a relatively straight line after the turn, and 2) adding a quadratic target velocity tracking cost. 
Lastly, it should be noted that the game is constructed such that the ego agent has no parameters in its own cost or constraints, so that it is only inferring the parameters of the simulated agents.}

\subsection{Experimental Setup}
\edit{For our first two examples, we aim to show that our method predicts trajectories more accurately than a baseline method, which solves the same inverse problem \emph{but presumes agents have a discount factor of $1$.}}\footnote{Full implementation details can be found at \url{https://github.com/cadearmstrxng/InverseGameDiscountFactor.jl}.}
Thus, the baseline estimator has fewer hidden parameters to estimate than our method.

\edit{To ensure a fair comparison, we use the same \acs{micp} back-end \cite{dirkse1995path} to satisfy the first-order necessary conditions of the \acs{golne} at line~\ref{line:solve-micp} of \cref{alg:1}, the same observation sequence, and the same initial guess---our method receives a uniformly random $\discountFactor \in [0,1]$ as an initial estimate.}
To compare the two methods, we calculate their respective error as $\error = \|\fullState -\fullState_{\textrm{ground-truth}}\|^2_2$,
enabling us to determine how well the methods predicted the agents' trajectories.
We gather these results with a Monte Carlo study done over a range of observation noise covariances \edit{as previously discussed}.
For full state experiments, the mean, $h(\state_t)$, is the identity function.
For partial state experiments, $h(\state_t) =[
p_{t}^{i\top} ~\forall i \in \numPlayers]^\top$.
The covariance is given by identity matrices scaled by 50 evenly spaced constants between 0 and \edit{\SI{0.01}{\meter\squared}}.
For each covariance matrix, we generated 50 noisy observation sequences given to our method and the baseline method, resulting in $2500$ observation sequences and parameter estimates.
\edit{Results are shown in \cref{fig:traj_error} and \cref{fig:Ind_error}, and discussed below.}

\begin{figure}
    \centering
    \includegraphics[width = 0.7\columnwidth]{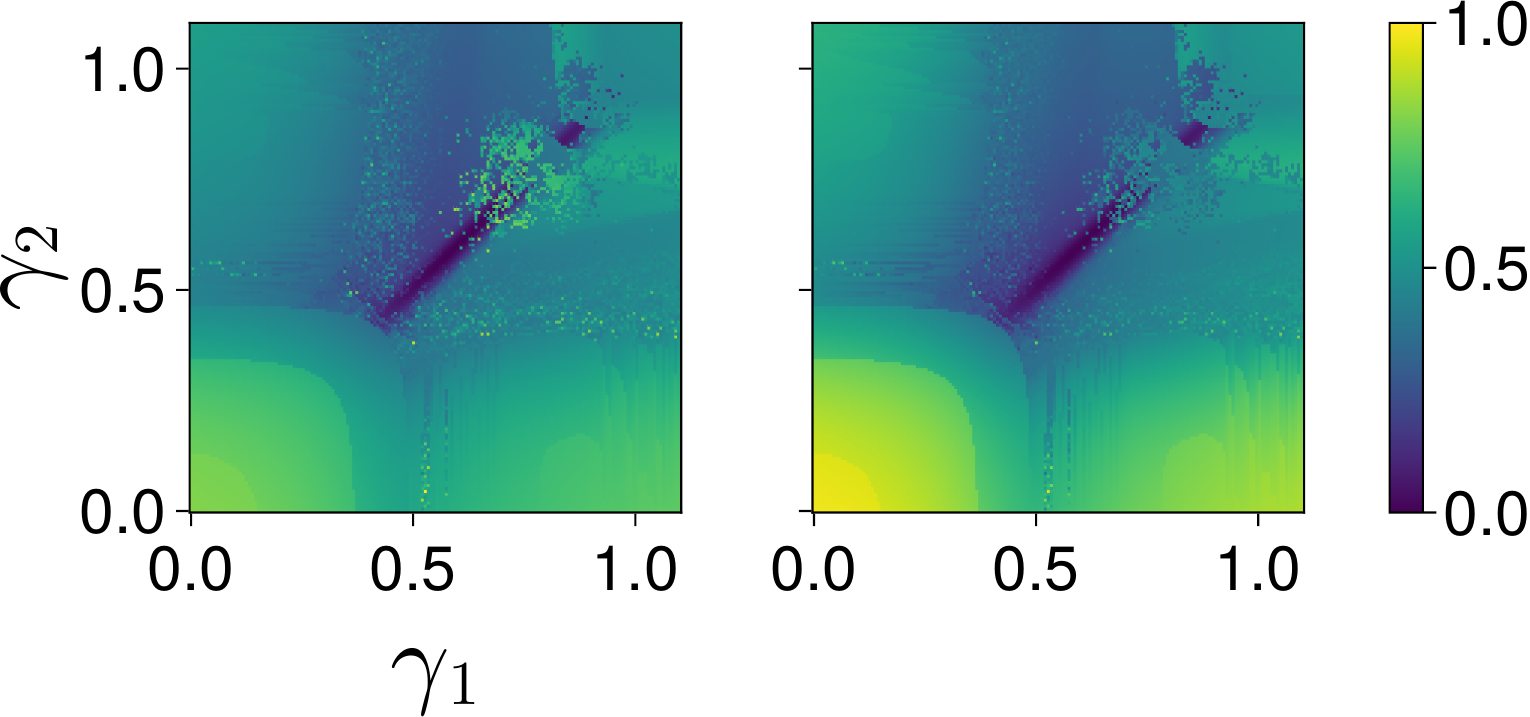}
    \vspace{-0.2cm}
    \caption{Heatmap of $\probMin$ from \cref{prb:InverseProblemStatement} \edit{for the crosswalk experiment for the fully observable (left) and partially observable (right) cases.
    Both players' $\discountFactor$ are varied while holding $\gameParam$ constant} at the ground truth values and the resulting cost is scaled to be in between $[0, 1]$.
    \vspace{-0.5cm}}
    \label{fig:heatmaps}
\end{figure}
\edit{
For the Waymax example, we aim to show our method can be used in a robotics context safely and efficiently by employing it in a receding horizon planner.
At each time $t$, the ego robot infers other agents' cost parameters from the past 10 observations and plans over the next 10 time steps.
The action used at $t$ is the first action taken in the planning trajectory.
To evaluate safety and efficiency, we simulate 50 random sequences of observation noise with standard deviation $\SI{0.1}{\meter}$ and plot the ego robot's distance to each other robot and to its goal position over the time horizon, shown in \cref{fig:waymax metrics}. 
Results are qualitatively similar at lower noise levels.
}
\subsection{Monte Carlo Results - Crosswalk}
\cref{fig:trial_image} shows a representative trial for our running example at a simulated crosswalk.
It shows the two players' observations, initial positions, ground-truth goals, inferred goals for the baseline and our method, and current and planned trajectories at a snapshot in time.
Note that our method results in a \edit{trajectory closer to the observations in L$_2$ distance than the baseline}.
In addition, according to the bar graph, our method accurately recovers the discount factor within $0.3\%$.

\cref{fig:heatmaps} shows a heatmap visualizing the objective function for \cref{prb:InverseProblemStatement} in which $\gameParam$ is fixed to the ground truth values, 
in both fully- and partially-observed settings. 
\edit{The diagonally-shaped low-cost region implies that the \textit{equilibrium} of the game is relatively insensitive to the  agents' absolute discount factors as long as both agents have similar discount factors (and both are greater than $\approx 0.5$).
We expect this low-cost region to exist because of the game's highly symmetric structure, motivating us to introduce a \emph{regularization} scheme for the inverse game in \cref{prb:InverseProblemStatement} to encourage better solver performance.
To do so, we added the term $\regConstant_\discountFactor \|1 - \discountFactor\|_2^2$ to \cref{eq:cov_distance_prob}, which encodes a prior belief that agents are more farsighted; it is straightforward to instead penalize $\|\discountFactor\|_2^2$ and encode the belief that agents are more shortsighted.
}
In the following Monte Carlo results, we report results for $\regConstant_\discountFactor = 10^{-3}$.
\edit{For clarity, we do not regularize in the other, asymmetric experiments.}

\begin{figure}
    \centering
    \subfigure[Only position measurements\label{fig:crosswalk_po_error}]{\includegraphics[width = 0.4\columnwidth]{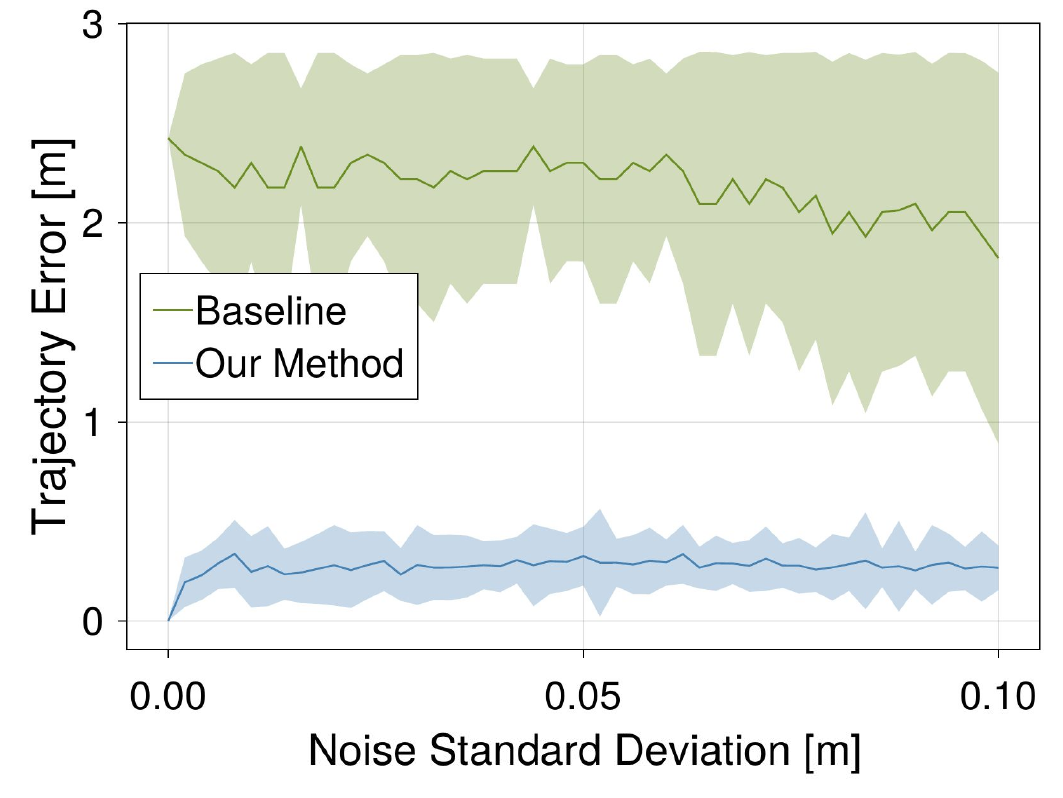}
    }
    \subfigure[Full state measurements\label{fig:crosswalk_fo_error}]{\includegraphics[width = 0.4\columnwidth]{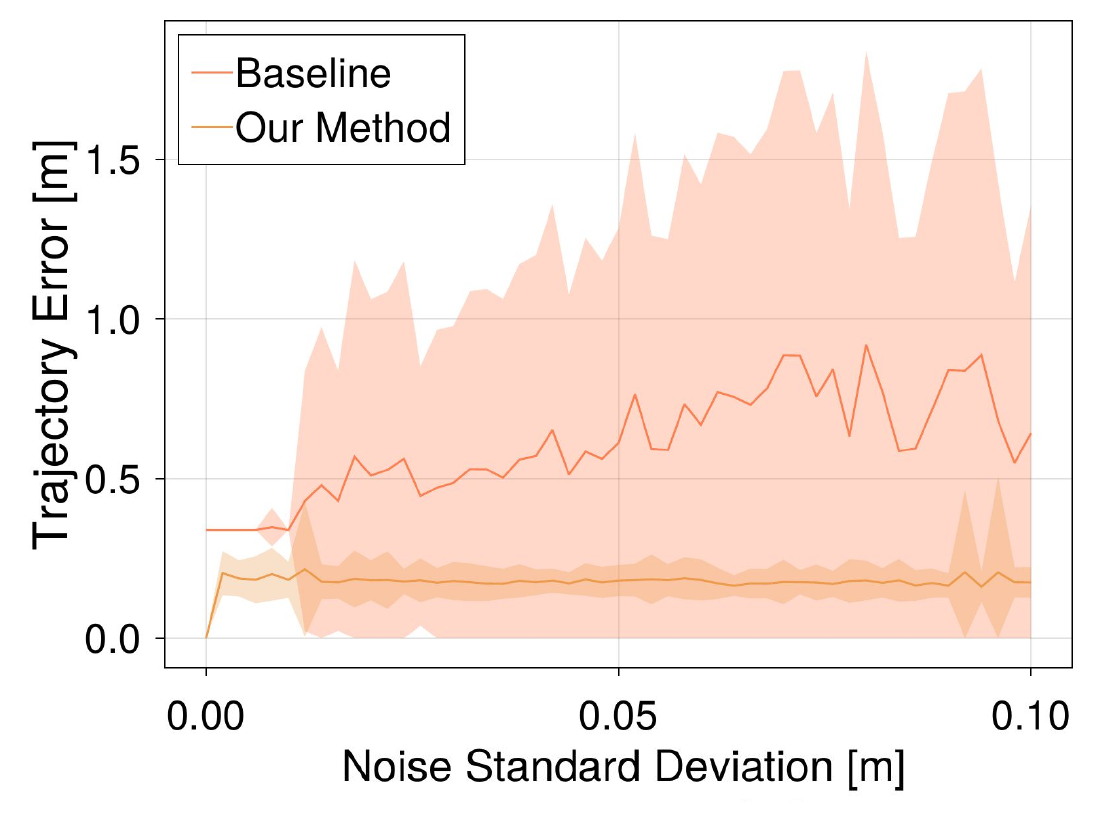}
    }
    \caption{\cref{alg:1} reliably \edit{improves predicted trajectory error in the crosswalk experiment, for partial and full state observations.}
    Solid/dotted lines denote means and the opaque band indicates standard deviation.
    \vspace{-0.5cm}}
    \label{fig:traj_error}
\end{figure}

\cref{fig:traj_error} shows the mean trajectory estimation error as a function of the observation noise standard deviation for \cref{alg:1} and the $\discountFactor = 1$ baseline.
Our method substantially outperforms the baseline \edit{in both cases.}
Interestingly, we can see that the mean error is only marginally sensitive to noise for both methods.
Rather, the most noticeable effect of noise is an increase in the standard deviation of the errors. 
From these results, we conclude that our foresighted formulation can better predict the behavior of short-sighted agents in comparison to a state-of-the-art inverse game baseline \cite{liu2023learningplaytrajectorygames}.

\subsection{Monte Carlo Results - InD Intersection}
\label{subsec:intersection_results}
As previously discussed, \cref{fig:InD_vis} shows a representative trial of our method and the baseline with the real-world data from the InD dataset \cite{inDdataset}. 
It shows the initial positions, the perceived and inferred goals, the observations and recovered trajectories up to a time $t$ = 15, and the planned trajectories for all four players at $t$ = 15 for the baseline and our method.
Our visualization shows a closer recovery of the ground-truth goal position than the baseline.
In addition, we recover a \edit{trajectory closer to the observations in L$_2$ distance than baseline.}

\begin{figure}
    \centering
    \subfigure[Only position measurements\label{fig:Ind_po_error}]{\includegraphics[width = 0.4\columnwidth]{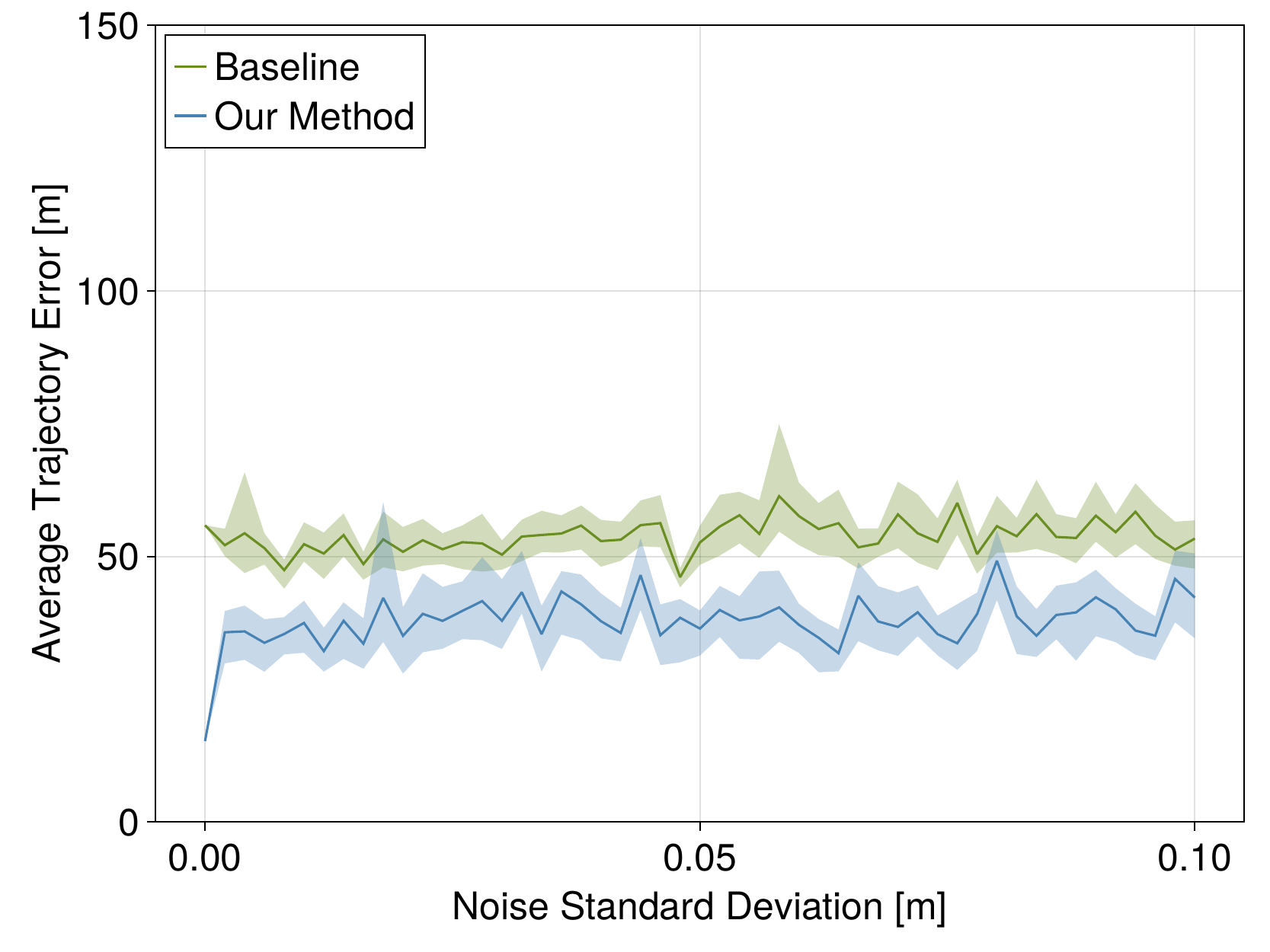}
    }
    \subfigure[Full state measurements\label{fig:Ind_fo_error}]{\includegraphics[width = 0.4\columnwidth]{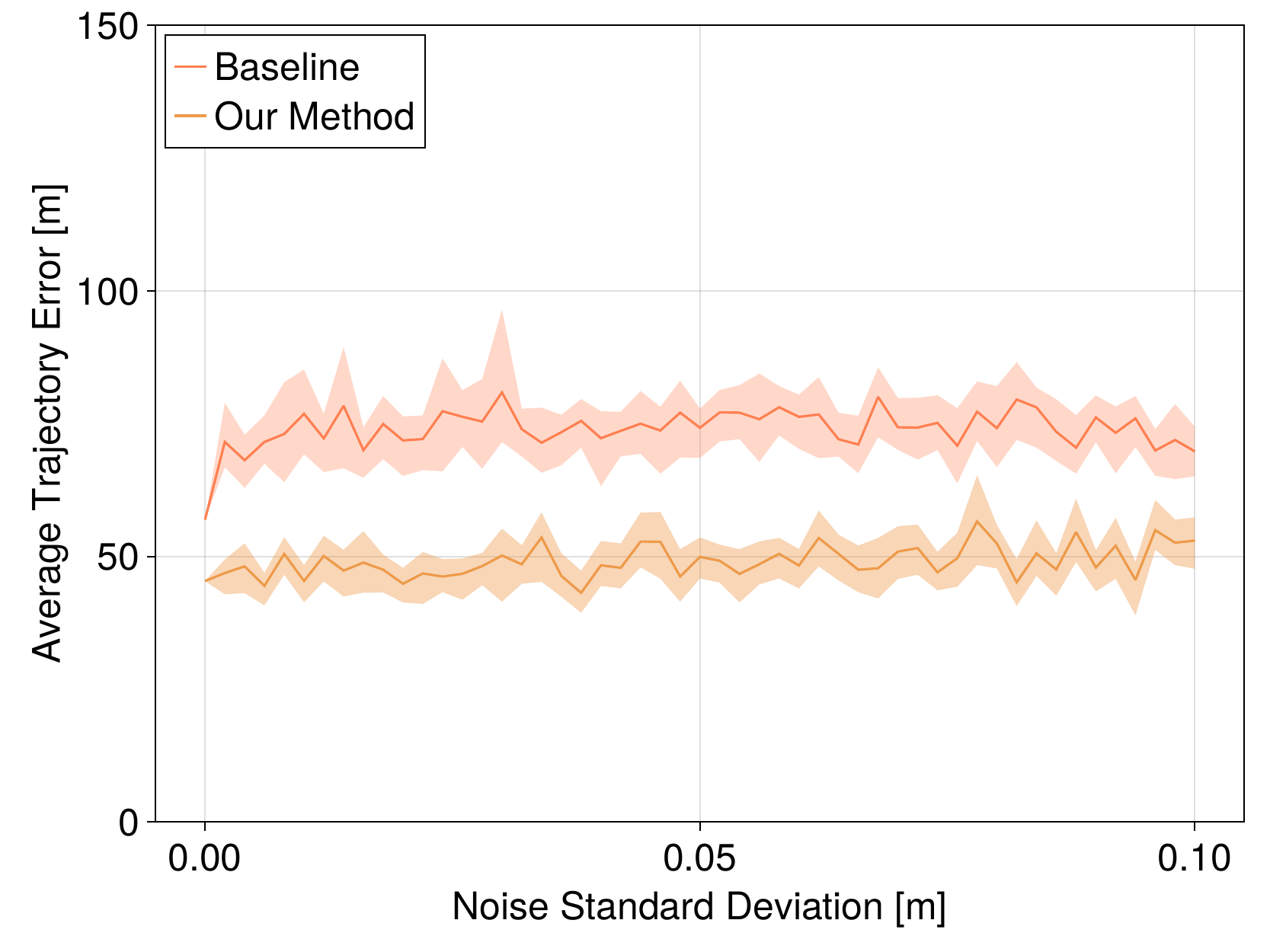}
    }
    \caption{\edit{
    Bootstrapped estimates (solid lines) and standard errors for the average trajectory prediction error for the baseline method and our method for the InD intersection experiment.
    On average, our method produces trajectories with less error in settings with both (a) partial state observations and (b) full state observations.
    }
    \vspace{-0.5cm}}
    \label{fig:Ind_error}
\end{figure}

\cref{fig:Ind_error} shows the mean trajectory error as a function of the observation noise standard deviation for the previously described methods. 
\edit{On average, our method consistently shows a significant improvement on the baseline in both the fully observable and partially observable cases.}
However, both methods seem to be relatively insensitive to increases in noise, as they create only a small increase in \edit{average} error.
Furthermore, unlike the synthetic data case, both methods are also relatively consistent in terms of error standard deviation.

\edit{
Note that in both the full and partial state experiments, the noiseless-observation trajectory error is non-zero.
We attribute these errors to model mismatch, referring to the differences between actual human decision-making processes and our recreation of human decision-making.
Two likely dominant sources of mismatch are our use of an open-loop information structure---chosen for tractability over a more human closed-loop form---and the expressiveness of our cost structures.
More complex cost structures, such as those encoded by neural networks, may also close the gap. 
}

\subsection{\edit{Monte Carlo Results - Waymax}}

\edit{
Finally, we show how our method can be integrated into a receding horizon motion planner, and present several metrics related to safety and efficiency.
In \cref{fig:waymax metrics}, we show that a foresighted game-theoretic receding horizon planner maintains a distance of over $\SI{6}{\meter}$ from all cars at all times across all noise realizations, exhibiting a high level of safety.
Meanwhile, the ego agent is reliably able to minimize the distance to its goal across all trials.
These results demonstrate that our foresighted formulation yields safe and efficient behavior when employed in receding horizon robotics schemes. 
A representative trial of this simulation is shown in \cref{fig:waymax}.
}

%% file: Content/07_DiscussionAndConclusions.tex
\section{Conclusions}
This paper formulates a noncooperative game that models potentially foresighted agents by associating each one with an unknown discount factor.
Our approach rewrites the equilibrium conditions of the resulting game as an \acs{micp} to leverage the directional differentiability of its equilibrium solutions with respect to unknown parameters.
This construction allows us to solve an \emph{inverse game} problem and identify hidden parameters---including agents' unknown discount factors---via a gradient descent procedure.
By evaluating our method with noisy partial and full state observations on a simulated crosswalk environment and a real-world intersection and road, we demonstrate superior performance compared to an existing state-of-the-art method. 

This work models foresighted behavior through the use of discount factors.
In this vein, we find several avenues for further research interesting.
\edit{First, because neural networks are universal function approximators, they can be used to approximate each player's cost function.
This could potentially eliminate model mismatch and show greater improvement on our method.
Second, humans often make decisions under the assumption they will have more information in the future, i.e., within feedback information structures.
As noted in \cref{subsec:intersection_results}, future work may extend our open-loop formulation to better model this behavior, with \citet{li2023cost} offering a clear starting point for this direction.
Third, humans may adapt how far ahead they plan based on context. 
On clear highways, they may focus on the near term, but in complex situations like merging or heavy traffic, they consider longer horizons. 
Future work may explore discounting schemes that vary with state and time.
}

\begin{figure}
    \centering
    \includegraphics[width=\columnwidth]{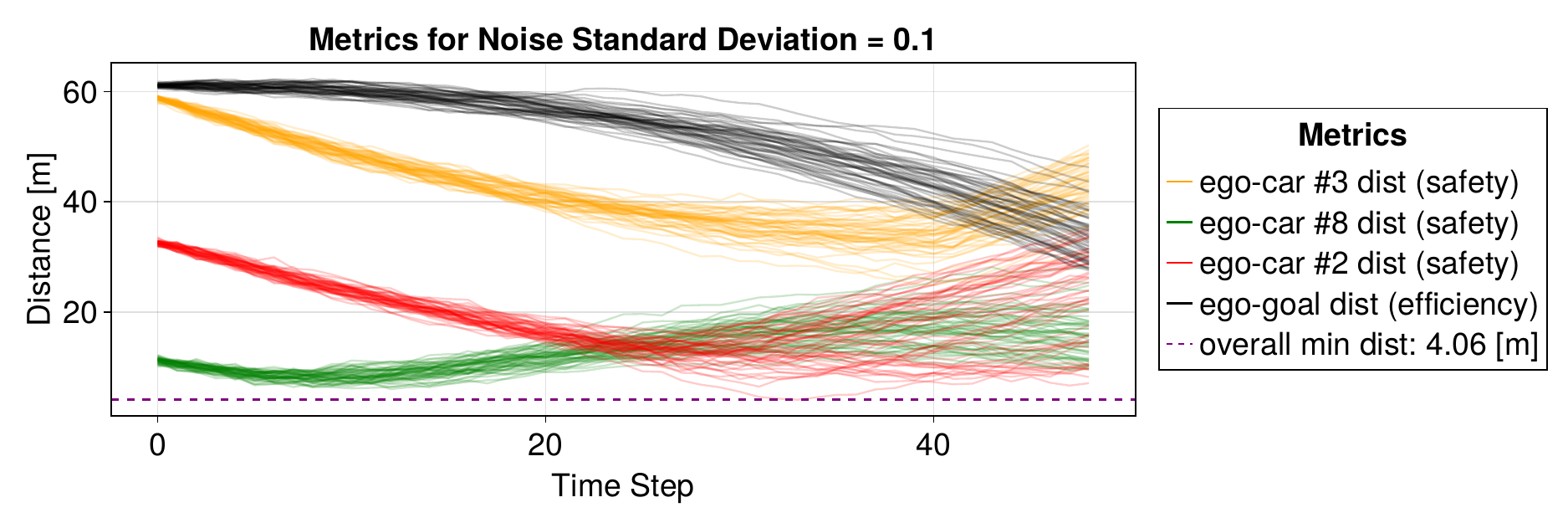}
    \caption{\edit{
    Several metrics computed on $50$ trajectories generated from the Waymax Monte Carlo simulation with an observation noise level of $\SI{0.1}{\meter\squared}$. In particular, we focus on the distance between the ego robot and the other most important drivers in the simulation, as well as how far the ego robot is from their goal.
    }
    \vspace{-0.5cm}
    }
    \label{fig:waymax metrics}
\end{figure}